%% file: main.tex
\documentclass[letterpaper]{article} 

\usepackage{aaai24}  
\usepackage{times}  
\usepackage{helvet}  
\usepackage{courier}  
\usepackage[hyphens]{url}  
\usepackage{graphicx} 
\usepackage{natbib}  
\usepackage{caption} 
\usepackage{algorithm}
\usepackage{algorithmic}
\usepackage{graphicx,subcaption}
\usepackage{adjustbox}
\usepackage{newfloat}
\usepackage{listings}

\DeclareCaptionStyle{ruled}{labelfont=normalfont,labelsep=colon,strut=off} 
\floatstyle{ruled}
\newfloat{listing}{tb}{lst}{}
\floatname{listing}{Listing}
\title{Interactive Visual Task Learning for Robots}
\author{
    Weiwei Gu,
    Anant Sah, 
    Nakul Gopalan
}
\affiliations{

    School of Computing and Augmented Intelligence, Arizona State University \\
    \{weiweigu, asah4, ng\}@asu.edu
}

\usepackage{booktabs}
\usepackage{xcolor}

\usepackage{enumitem}

\usepackage{bibentry}

\begin{document}

\maketitle

\begin{abstract}
We present a framework for robots to learn novel visual concepts and tasks via in-situ linguistic interactions with human users. Previous approaches have either used large pre-trained visual models to infer novel objects zero-shot, or added novel concepts along with their attributes and representations to a concept hierarchy. We extend the approaches that focus on learning visual concept hierarchies by enabling them to learn novel concepts and solve unseen robotics tasks with them.  To enable a visual concept learner to solve robotics tasks one-shot, we developed two distinct techniques. 
Firstly, we propose a novel approach, Hi-Viscont(HIerarchical VISual CONcept learner for Task), which augments information of a novel concept to its parent nodes within a concept hierarchy. 
This information propagation allows all concepts in a hierarchy to update as novel concepts are taught in a continual learning setting. 
Secondly, we represent a visual task as a scene graph with language annotations, allowing us to create novel permutations of a demonstrated task zero-shot in-situ. 
We present two sets of results. 
Firstly, we compare Hi-Viscont with the baseline model (FALCON~\cite{mei2022falcon}) on visual question answering(VQA) in three domains. 
While being comparable to the baseline model on leaf level concepts, Hi-Viscont achieves an improvement of over $9\%$ on non-leaf concepts on average.
Secondly, we conduct a human-subjects experiment where users teach our robot visual tasks in-situ. We compare our model’s performance against the baseline FALCON model. 
Our framework achieves $33\%$ improvements in success rate metric, and $19\%$ improvements in the object level accuracy compared to the baseline model.
With both of these results we demonstrate the ability of our model to learn tasks and concepts in a continual learning setting on the robot. 
\end{abstract}

\section{Introduction}
\input{sections/introduction}

\section{Related Work}
\input{sections/related_works}

\section{Methods}
\input{sections/methods}

\section{Results}

\input{sections/results}

\input{sections/discussions}

\bibliography{custom}

\newpage
\hfill

\newpage
\appendix
\section{Implementation Details}\label{app:training_details}
\input{appendices/training_details}

\section{Dataset Statistics}\label{app:dataset_stats}
\input{appendices/dataset_statistics}

\section{Detailed Results}\label{app:detailed_results}
\input{appendices/detailed_results}

\end{document}

%% file: sections/introduction.tex


\begin{figure*}[h]
  \centering
\begin{subfigure}{0.28\textwidth}
\includegraphics[width=\textwidth,page=1]{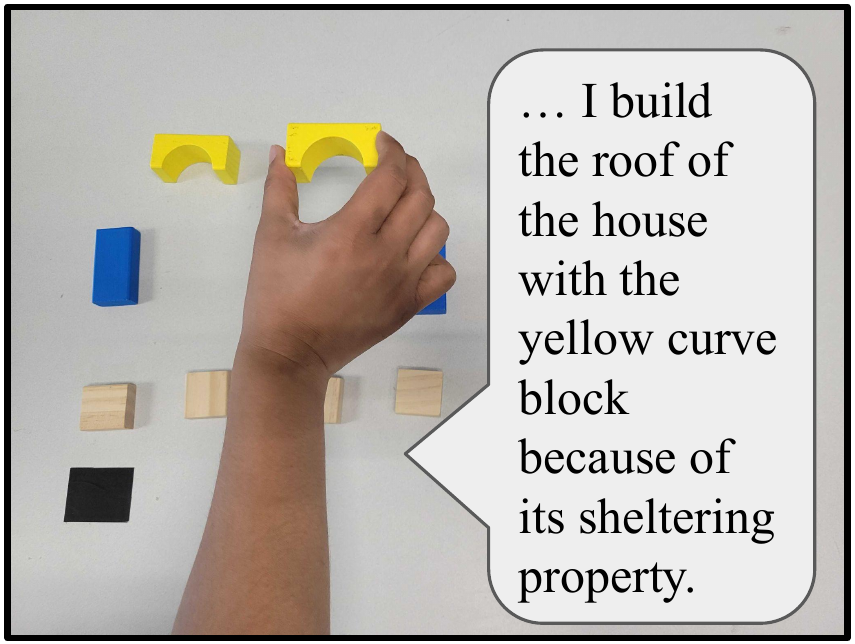}
  \subcaption{}
\end{subfigure}
\hspace{0.01\textwidth}
\begin{subfigure}{0.28\textwidth}
\includegraphics[width=\textwidth,page=2]{figures/figure1_breakdown.pdf}
  \subcaption{}
\end{subfigure}
\hspace{0.01\textwidth}
\begin{subfigure}{0.28\textwidth}
\includegraphics[width=\textwidth,page=3]{figures/figure1_breakdown.pdf}
  \subcaption{}
\end{subfigure}

\caption{This figure demonstrates how Hi-Viscont learns from users interactively. (a) First, the user demonstrates a structure, say a ``house,'' with its sub-components such as a ``roof'' and the concepts used to make the ``roof'' such as a ``yellow curve block''. (b) The user then teaches a novel concept such as a ``green curve block'' and describes its properties. (c) The user can now ask the robot to create a new structure (``house with green roof'') zero-shot with the taught component without explicitly asking for the object of interest. }
\label{fig:interactive_task_teaching}
\end{figure*}

Robots in a household will encounter novel objects and tasks all the time. For example, a robot might need to use a novel vegetable peeler to peel potatoes even though it has never seen, let alone used such a peeler before. Our work focuses on teaching robots novel concepts and tasks one-shot via human-robot interactions,
which include demonstrations and linguistic explanations.
We then want the robot to generalize to a similar but unseen visual task. 
A robotic system that can learn generalizable tasks and concepts from few natural interactions from a human-teacher would represent a large leap for robotics applications in everyday settings. 
In this work we aim to take a step in the direction of generalizable interactive learning as demonstrated Fig.~\ref{fig:interactive_task_teaching}. 

Previously, large image and language models have been extended to robotics to manipulate novel objects, and create visual scenes~\citep{shridhar2021cliport, brohan2023rt2}. 
These methods recognize novel objects by using their underlying large language and visual models to extract task-relevant knowledge.
However, they are not capable of learning to create a novel visual scene from  in-situ interactions with a human user.
There is also significant work in few-shot learning of  visual concepts in computer vision~\citep{mei2022falcon, snell2017prototypical,vinyals2017matching, sung2018learning, wang2018zeroshot, tian2020rethinking}, albeit without extensions to robotics domains.
These approaches focus on learning novel concepts for image classification, but ignore the fact that the novel concepts also bring new information to update our understanding of concepts already known to the robot.
The reverse path of knowledge propagation, that is, from novel concepts to previously known concepts is equivalently important in performing tasks in the real-life scenarios, especially when the agent has little knowledge of the world and needs to continually add information to known concepts. 

In this work, we propose a novel framework, Hi-Viscont, that enables robots to learn visual tasks and visual concepts from natural interactions with a human user. We learn the task type and concepts from users one-shot, and then generalize to novel task variants within the task type zero-shot.  
We do this by connecting our insights on \emph{one-shot visual concept learning} and the use of \emph{scene graphs}. 
The robot learns the structure of a visual task by converting linguistic interactions with a human user into a scene graph with language annotations. 
Moreover, Hi-Viscont updates parental concepts of the novel concept being taught. Such updates allow us to generalize the use of the novel concepts in to solve novel tasks.


%
The contribution of this work is three-fold:
\begin{enumerate}[noitemsep,topsep=0pt,parsep=0pt,partopsep=0pt]
    \item We present concept learning results on VQA tasks that are comparable to the state-of-the-art FALCON model. More specifically, Hi-Viscont improves on FALCON on all non-leaf concepts across all domains with significance.
    \item We enable the robot agent to learn a visual task from in-situ interactions with a scene graph, allowing zero-shot generalization to an unseen task of the same type, as demonstrated in Fig.~\ref{fig:interactive_task_teaching}. 
    \item Finally, we conduct a human-subjects experiment to show that our system is able to learn visual tasks and concepts from in-situ interactions with human users. Hi-Viscont achieves a $33.33\%$ improvement in Success Rate when completing the users' requests compared to FALCON ($p=0.014$).
    
\end{enumerate}

%% file: sections/related_works.tex
\textbf{Language conditioned manipulation.} 
Significant work has been performed in  learning concepts and tasks for robots in interactive settings~\cite{gopalan2018sequence,gopalan2020simultaneously,tellex2020robots} even with the use of dialog~\cite{chai2018language,matuszek2012joint}.
Our work differs from previous works as it learns visual concepts for manipulation one-shot, and improves generalization by updating other known concepts.
Moreover, our approach can learn a concept hierarchy starting from zero known concepts, displaying the adaptability of our model under a continual learning setup.
Previous work has focused on language conditioned manipulation~\citep{shridhar2021cliport, liu2021structformer, brohan2023rt1, brohan2023rt2}. \citealt{shridhar2021cliport} computes a pick and place location conditioned on linguistic and visual inputs. 
\citealt{liu2021structformer} focuses on semantic arrangement on unseen objects. 
Other works train on large scale linguistic and visual data and can perform real-life robotic task based on language instructions~\citep{ahn2022i, brohan2023rt1, brohan2023rt2}. Our work focuses on interactive teaching of tasks and concepts instead of focusing on the emergent behaviors from large models. 
\citealt{daruna2019robocse} learns a representation of a knowledge graph by predicting directed relations between objects allowing a robot to predict object locations. 
To the best of the author's knowledge, our work  is the first that learns concepts and tasks one-shot to generalize to novel task scenarios on a robot, making our contributions significant compared to other related works.

\noindent\textbf{Visual reasoning and visual concept learning.} Our work is related to visual concept learning \citep{mei2022falcon, mao2018the, yi2019neuralsymbolic, han2020visual, li2020competenceaware} and visual reasoning \citep{Mascharka_2018, DBLP:journals/corr/abs-1807-08556, DBLP:journals/corr/JohnsonHMHLZG17, DBLP:journals/corr/abs-1803-03067}. To perform the visual reasoning task, traditional methods \citep{Mascharka_2018, DBLP:journals/corr/abs-1807-08556, DBLP:journals/corr/JohnsonHMHLZG17, DBLP:journals/corr/abs-1803-03067} decompose the visual reasoning task into visual feature extraction and reasoning by parsing the queries into executable neuro-symbolic programs.  On top of that, many concept learning frameworks \citep{mei2022falcon, mao2018the, yi2019neuralsymbolic, han2020visual, li2020competenceaware} learn the representation of concepts by aligning concepts onto objects in the visual scene. 
As far as we know, \citealt{mei2022falcon}'s FALCON is the most similar work to our work in this line of research. However, when introducing a new concept, our work continually updates the representation of all related concepts, whereas \citealt{mei2022falcon} does not, which makes it ill-suited for continual learning settings. Our work is also related to the area of few-shot learning~\citep{snell2017prototypical, tian2020rethinking, vinyals2017matching}, which learns to recognize new objects or classes from only a few examples but does not represent a concept hierarchy which is useful in robotics settings.



\noindent\textbf{Scene graph.} Scene graphs are  structural representations of all objects and their relationships within an image. The scene graph representation~\cite{Chang_2023} of images is widely used in the visual domains for various tasks, such as image retrieval\cite{DBLP:journals/corr/JohnsonHMHLZG17}, image generation\citep{johnson2018image}, and question answering\citep{teney2017graphstructured}.
This form of representation has also been used in the robotics domains 
for long-horizon manipulation~\citep{zhu2021hierarchical}.

%% file: sections/methods.tex
\begin{figure*}
    \centering
    \includegraphics[width=0.75\textwidth]{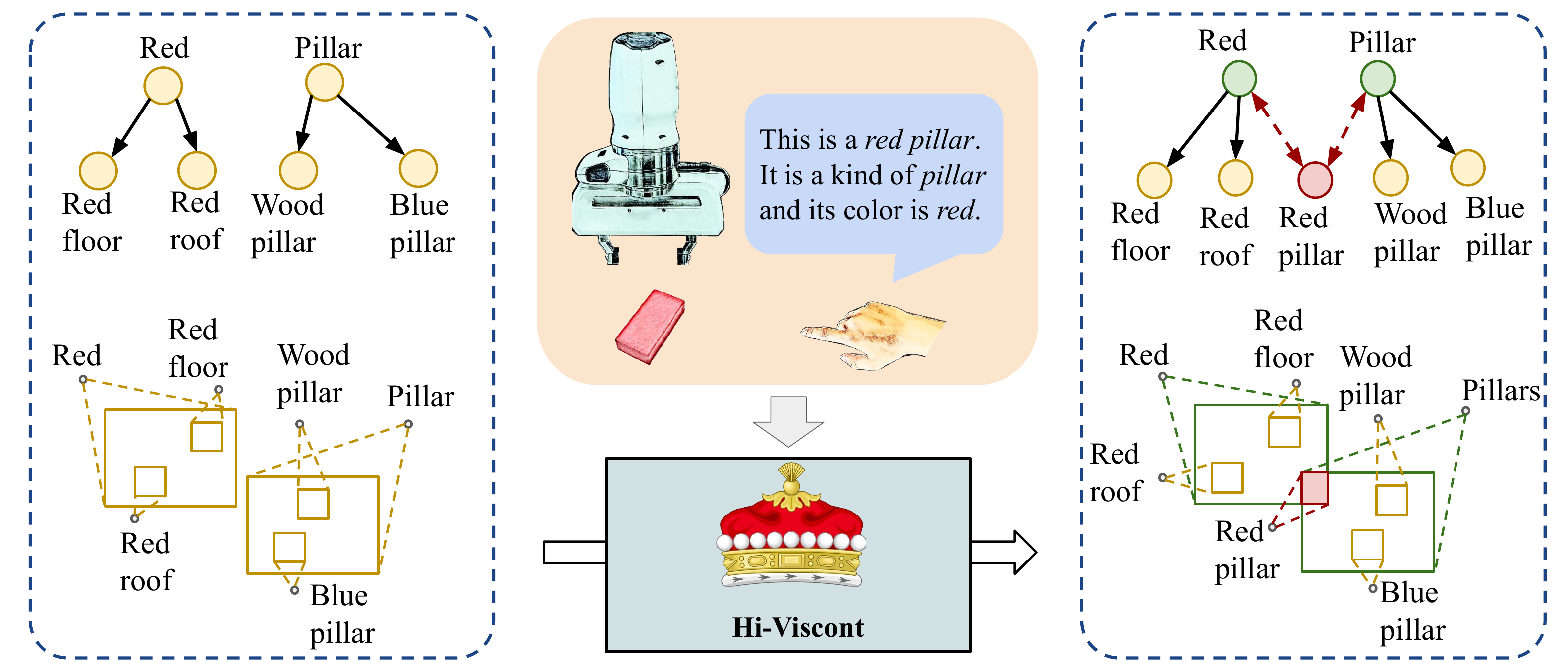}
    \caption{We demonstrate the updates to the box embedding space and the parent concepts when a novel concept is taught to our robot using Hi-Viscont. Existing approaches only edit the leaf nodes as those represent novel concepts.}
    \label{fig:concept_net_model}
\end{figure*}

We first present the baseline FALCON model and then introduce our Hi-Viscont model. Our model is based on concept learners as they learn concepts few shot, and can reason over the attributes of chosen (and their parent) concept classes. FALCON is a State-of-the-art (SOTA) concept learner which learns novel concepts one-shot. 

\subsection{FALCON}
\citet{mei2022falcon} developed FALCON, a meta-learning framework for one-shot concept learning in visual domains.
FALCON learns a new visual concept with one or a few examples, and uses the learned concept to answer visual reasoning questions on unseen images.
There are three components for the model: a visual feature extractor that extracts the object-centric features for the input image, a graph neural network (GNN) based concept learner, and a neuro-symbolic program executor that executes the input neuro-symbolic program.

Natural language sentences describing objects and their queries are represented as structured neuro-symbolic programs.
FALCON learns novel concepts by interpreting the images presented and the relationships between known concepts and the unknown concept being learned using a neuro-symbolic program. 
After learning, the model performs reasoning over questions, executed as neuro-symbolic programs to answer questions about images. 

A pre-trained ResNet-34 is used as visual feature extractor for the model.
The visual feature extractor computes a feature for each object in a scene separately, which can be used for downstream visual reasoning.
FALCON uses a box embedding\citep{vilnis2018probabilistic} to represent concepts and their object visual features.

Finally, the concept learning module is composed of two separate Graph Neural Networks(GNNs), the Relation GNN and the Example GNN.
To compute a representation for a novel concept $c$, FALCON starts with an embedding that is randomly sampled from a prior Dirichlet distribution.
Then, the model updates the representation of $c$ by computing messages from parent nodes based on their factor weights or relationship and also computing a message from the visual feature (represented as a node within the Example GNN) for the concept being learned.  
This representation for the novel concept $c$ will then be used for downstream VQA tasks. 
There are two major issues to directly use FALCON for interactive task learning on the robot. Firstly, the model lacks scene information to solve tasks. We address this in our work. Secondly, the model assumes concepts are learned perfectly and do not need to be updated as it learns more concepts.
For example, when we teach the model the concept of ``container'' with an image of a ``cup,'' FALCON cannot update the features of the ``container'' concept when the concept of ``bowl'' is taught as a child to the ``container'' concepts. This allows FALCON to learn that all ``containers'' have handles which is untrue.
\begin{figure*}[h]
    \centering
    \includegraphics[width=0.75\textwidth]{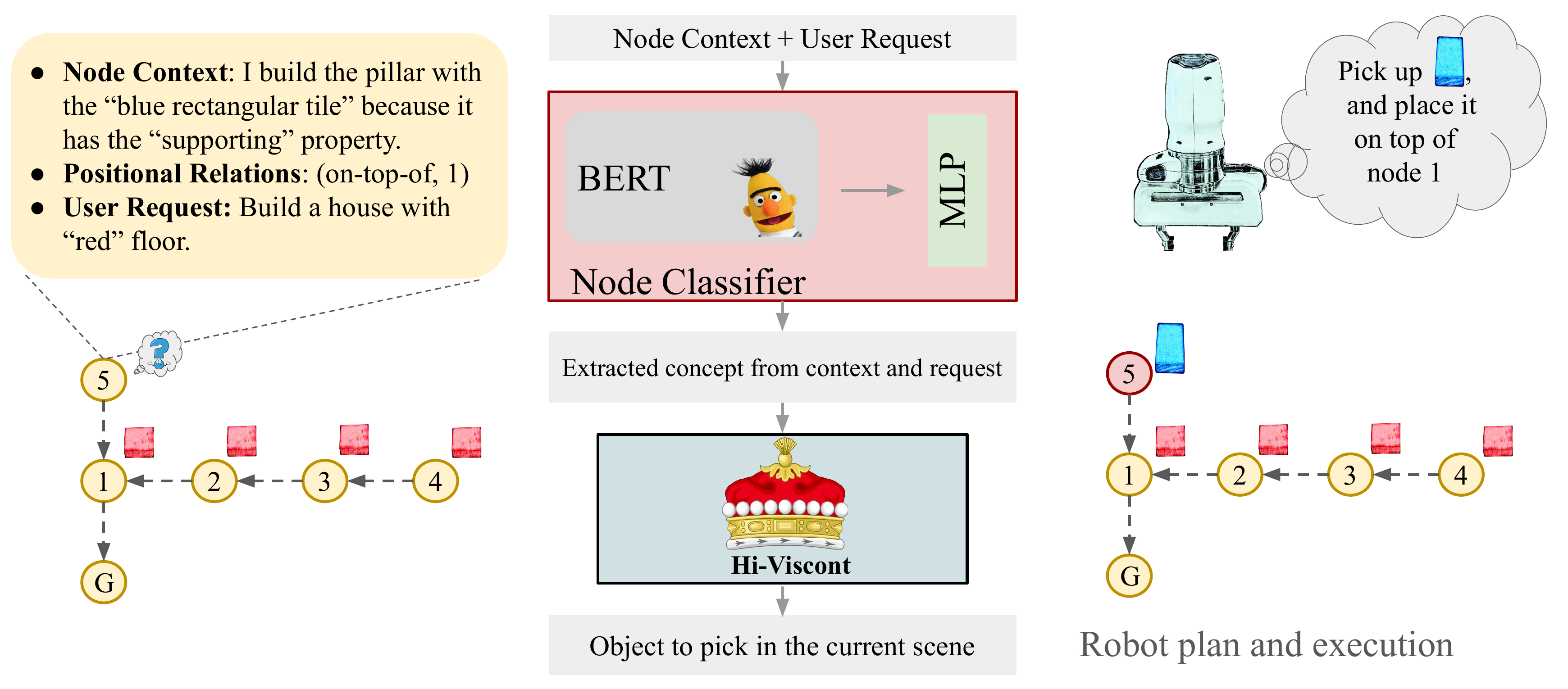}
    \caption{Our pipeline decides objects for each node in the scene graph one at a time. The node's context and the request phrase are fed into a node classifier, which is composed of a BERT encoder and an multilayer perceptron, to decide the concepts applicable for the current node. Hi-Viscont then decides the object to pick in the current scene based on the extracted concepts. In this example, the object chosen for Node 5 is "blue rectangular tile" as it existed in the original demonstration and was not changed given the novel task's linguistic request. Notice that the new structure has red floor tiles which were never demonstrated to the robot.}
    \label{fig:node_classifier}
\end{figure*}

\subsection{Hi-Viscont}
We present our concept net model, Hi-Viscont (HIerarchical VISual CONcept learner for Task), which actively updates the related known concepts when we introduce the novel concept to improve upon FALCON's generalization capabilities.
We adopted several modules from the framework of FALCON, including the visual feature extractor, the neuro-symbolic program executor, the box embedding space, and the novel concept learner.
Moreover, we introduce an additional GNN module, Ancestor Relational GNN (ARGNN), that updates the related known concepts as a novel concept is introduced. 
ARGNN predicts a new embedding for the related known ancestor concepts to the novel concept. To do this update we compute a message from the visual feature of novel concept's instance to the embedding of the related nodes using the relations between the parent concepts and the novel concept.

When a novel concept $c$ is inserted to Hi-Viscont, the extracted visual feature $o_c$ of concept $c$ and its relations with known concepts $R_c$ are fed to Hi-Viscont as input. 
Each relation $rel=(c',c, r)$, where $c'$ denotes the related concept, and $r$ describes its relationship with $c$.
We compute an embedding $e_c$ for novel concept $c$ using the same method as FALCON.
Using the additional ARGNN, we predict a new embedding for each related concept $c'$ by computing a message from the visual feature $o_c$ to the current embedding of the related concept $e_{c'}^0$ using the same relationship $rel$.
The formula for this update is denoted as follows: $$e_{c'}^1 = \textrm{ARGNN}(o_c, rel, e_{c'}^0)$$
The resulted embedding $e_{c'}^1$ will be used as the representation for concept $c'$ for future task or updates.

To provide gradient flow to train ARGNN, we extended the concept learning task proposed by FALCON by adding validation questions for each related concept, that is when a new concept is added all concepts in the concept net are tested for accuracy over the novel concept. For example, from our previous discussion the newly inserted ``bowl'' concept's object instance is checked with the ``container'' parent to see if the presented ``bowl'' also tests as a ``container.''   A more detailed description of our training pipeline and  methodology can be found in the appendix.
While FALCON was evaluated solely on the newly inserted concept, we evaluate all concepts (leaf and parent nodes) of our model on unseen images. Such an evaluation ensures consistency between parent and child concepts which is a necessity in continual learning settings. 
This evaluation mechanism allows us to evaluate the quality for the embedding of all concepts in the resulting knowledge graph, which is closer to how these knowledge are used in the real world setting.

\subsection{Learning Visual Task via Scene Graph}
Figure~\ref{fig:interactive_task_teaching} illustrates how our pipeline learns a visual task from a single in-situ interaction with a human user.
The user's demonstration(Fig.~\ref{fig:interactive_task_teaching}.a) is first converted  into an initial scene graph.
Each node of the initial scene graph corresponds to an object that the user placed, and it contains the bounding box information of the object and the user's linguistic description of the object.
We also store the positional relations with respect to other nodes for each node, which allows for object placements when reconstructing the scene.
A fixed location on the table is marked with black tape as the origin, which is treated as the zeroth object. All other objects placed by the user will be to the top right of the origin.
%

Based on the initial scene graph and the user's linguistic request for the desired variant of the visual scene, we infer a goal scene graph modelled as a node-wise classification task as shown in Figure~\ref{fig:node_classifier}.
Since the variant of the visual task from the user request shares the same structure as the demonstration, the goal scene graph has the same number of nodes as the initial scene graph.
We take the user's description of the corresponding node of the initial scene graph $t_i$ and the user's request of the variant of the structure $q$ as inputs, and perform a two-step inference: First we decide if the node in the goal graph is different from the one in demonstration; Subsequently if the node is different we decide with classification which object satisfies the node location.

To decide whether the concept of a node within the scene has changed given the user's description of the node and the user's current request $q$, we perform a binary classification at each node. The result of this classification decides if we are changing a node's concept or not. 
We use a pretrained BERT\textsubscript{base} model to encode the context request pair, which is then fed into a multi-layer perceptron (MLP) with a Cross-Entropy loss.
The second step of the inference extracts the related concepts from the context if the node's concept needs to be changed as per the request.
We convert the concept extraction problem into a classification problem by providing concept candidates as a part of the input again with BERT model and an MLP with a Cross-Entropy loss.
The related concepts of each node are fed as input for the concept net model to decide the object to pick, and the positional relations with other nodes are used to compute the placement location.
The robot reconstructs the scene following the order of the nodes.
Pairing the concept net model with scene graph, the robot is able to learn the arrangement of a scene in one single demonstration and perform variants of the scene without demonstration. We allow FALCON access to our Scene Graph classifiers to have a valid baseline to compare against.

\subsection{Robotics Setup for User Study}
We integrate our visual task learning and concept learning model with a Franka Emika Resarch 3 arm (FR3). 
This pipeline allows us to show the generalizability with which Hi-Viscont learns visual concepts when compared to Falcon~\cite{mei2022falcon} in learning and solving novel tasks. 
To set this demonstration up we use a Franka Emika Research 3 arm (FR3), two calibrated realsense D435 depth cameras, and a mono-colored table to allow for background subtraction. We use the SAM (Segment Anything Model)~\cite{kirillov2023segment} to separate the foreground and the background and get individual bounding boxes for each of the blocks on the table. 
For pick and place, we initially experimented with Transporter networks~\cite{zeng2021transporter}
but finally used a simpler Visuo-Motor Servoing mechanism for reliability.
We expected users to maintain about an inch of space between each object in the scene to allow the robot to pick objects without collisions and for SAM to segment objects from the background accurately.
In the process of picking and placing, the robot autonomously recovers if an error is made.
Once an object is grasped, we place it into the Task scene based on the position calculated relatively  with respect to the previously placed object nodes or the zeroth origin object. This process is done iteratively until the whole scene graph is completed.

\subsection{Human-Subjects Experiment}
\subsubsection{Study Design}
We conduct a 1 $\times$ 2 within-subject experiment to measure the framework's ability to learn visual task and visual concepts from in-situ interaction.
We extend the FALCON model with our scene graph module and use it as a baseline to compare against because we could not find any equivalent prior work.
Both concept net models are trained with the same split of concepts and the same training data for the same number of steps.
Through this experiment, we aim to demonstrate that our framework achieves better performance than FALCON model because of the continual update for the known knowledge.
Participants for our experiment interact with the robot in three phases.
For each interactive phase, the participants only interact with the robot once, and the interaction is recorded as the input for both systems.
After the interactive phases, the participant observes the two systems construct the scene requested by the participant. Half the participants observe FALCON first and other half observe Hi-Viscont system first to avoid ordering confounds.

\subsubsection{Human Subjects Experiment Domain}
We evaluate our approach with the human-subjects experiment in a 2-D object rearrangement domain, which is a problem commonly used in language grounding and HRI research~\citep{liu2021structformer, shridhar2021cliport}.
The domain we choose for this study is the House-Construction domain which we introduce in Domains Section. We designed this domain as the users have the ability to create complex types of structures with different object classes.
Building blocks from children's toys were used as objects in this domain as they are varied and easy to grasp by the robot, as grasping is not a focus of our work.

\subsubsection{Metrics}
The objective metrics we collect for the human-subjects experiment are as follows. 
We measure the success rate (SR) of completing the user's request with complete accuracy, and the node level accuracy of each scene graph for both systems.
Both metrics are used to measure each system's ability to actually complete the visual task objectively.
The success rate metric gives us the insight of system's ability of completing the whole task, while the node level accuracy metric provides a more fine-grained result on few-shot object recognition.
In the post-study survey for each system, we administer the Perceived Intelligence and Anthropomorphism sub-scales of the Godspeed Questionnaire Series~\cite{Bartneck2009MeasurementIF}, Trust in Automated systems questionnaire~\cite{trust_survey}, System Usability Scale (SUS)\cite{article}.
In addition, we hand-crafted a direct comparison survey for preference between Hi-Viscont and the FALCON model.

\subsubsection{Procedure}
This study was approved by our university's Institutional Review Board (IRB). We recruited all participants through on campus advertisements. 
The study took under $90$ minutes with voluntary participation. The participants were not compensated for their efforts. The procedure of the study is as follows. 
Participants first fill out consent form and then the pre-study survey.
After the pre-study survey, we hand out a general introduction for the experiment of the study.
Then, we guide the participant through the task teaching phase, the concept teaching phase, and the request phase sequentially as described below. 
Before each phase, we provide a demonstration video and the instructions of the corresponding phase to the participants. The instruction videos is demonstrating a different task (bridge making) with different objects (foam blocks) than the actual task the participant is teaching. 
The anonymized instruction manual and the link to these videos are provided in the Appendix and the associated webpage~\footnote{\url{https://sites.google.com/view/ivtl}}.


\noindent\textbf{Task Teaching Phase - } In the task teaching phase, the participants teach the robot a visual task by demonstrating the scene with its constituent structures one by one. The participants also describe the structures and the objects used to build the structure with natural language. For example, a participant might build a house, with floors, pillars and a roof. While building the roof the participant might say ``This a roof which I build with the curved blue tile because of it's sheltering capability.'' The participants demonstrate each of the structures with their chosen language commands one after another to build a house. We record all descriptions in audio and convert them into text using audio to text tools.  

\noindent\textbf{Concept Teaching Phase - } In this phase, the participants teach a novel concept of their choice to both systems by showing the object to the camera, and describing the concept's properties, such as the color and functional characteristics of the object, in natural language. 
The description to the novel object concept is converted to neuro-symbolic programs, which are given to both models for updates as described in the Methods section.

\noindent\textbf{Request Phase - } In the request phase, the participants are asked to provide a request in natural language for a novel scene that they did not demonstrate in the task teaching phase. They are instructed to use the object taught in the concept teaching phase in the request. The requested task still needs to be a house which is the same task type as the demonstration, but not a house that the models have seen.

After completing the three phases, the participants watch the two systems construct their requested scene in real time with a robot in a randomized order. As each system finishes the construction, the participants are asked to fill a post-survey for that system. After both systems finish the construction, the participants also fill out a comparative survey.

\subsubsection{Hypotheses}
We have the following hypotheses:

\noindent\textbf{Hypothesis 1 - } Our framework will have a higher success rate in completing the user's request without any errors. We hypothesize that our framework will have a higher success rate than FALCON model because of its update to the related known concepts, which could be used in requests that point to the same object indirectly.

\noindent\textbf{Hypothesis 2 - } Our framework will have a higher node level accuracy. We hypothesize that our framework will achieve a higher accuracy in the node level than the FALCON model as Hi-Viscont can correctly predict the object in a node without direct queries specifying the type of object. FALCON  does not handle these indirect queries well as it does not update parent concepts with knowledge about novel leaf level concepts.

\noindent\textbf{Hypothesis 3 - } We hypothesize that our framework will achieve higher ratings on subjective metrics compared to the baseline because of its higher accuracy and competency in completing the requests.

%% file: sections/results.tex
\begin{table}[t]
\resizebox{\columnwidth}{!}{%
\begin{tabular}{c|ccc}
\toprule
Method    & CUB-200-2011            & House Construction     & Zoo           \\
\midrule
Hi-Viscont & 74.39$\pm$7.04 & 86.41$\pm$5.28& 83.50$\pm$8.44\\
FALCON    & 73.40$\pm$5.77 & 87.17$\pm$4.17& 85.12$\pm$6.64\\
\bottomrule
\end{tabular}%
}
\caption{The average F1 score and standard deviation of Hi-Viscont and FALCON on the test concepts across all the three domains.}
\label{tab:test_three}
\end{table}

In this section, we present two sets of results.
We first present experiment results on concept learning on the visual question answering task on three different domains.
The experiment results demonstrate that our concept net model learns better representation for concepts than the baseline model, and is more robust for continual learning across all domains.
Secondly, we present the results for the human-subjects experiments.
The results for the human-subjects study demonstrate that our framework Hi-Viscont can learn visual task via an in-situ interaction with human user with more accuracy and usability than the baseline model. 

\subsection{VQA Experiment Setup}
\paragraph{Domains.} We first present experimental results on VQA tasks for three domains: the \textbf{CUB-200-2011 dataset}, a \textbf{custom house-construction domain} with building blocks, and a \textbf{custom zoo domain} with terrestrial and aquatic animals. 
We created the House-Construction and Zoo domains because they allow us to construct arbitrarily hard tasks with different types of objects that a robot can grasp.
For each domain, we introduce additional general concepts on top of the existing concept classes to construct a concept hierarchy. The detailed descriptions and the statistics of the datasets can be found in the Appendix.

\noindent\textbf{Data Creation Protocol.} Following FALCON's data creation protocol, we procedurely generate training and testing examples for each domain. 
We generate descriptive sentences and questions based on the ground truth annotations of images and external knowledge, which is the relationship between concepts.
For all the descriptive sentences and the questions, we also generate the corresponding neural-symbolic programs.

\noindent\textbf{Experiment Configuration.}
We directly compare Hi-Viscont with FALCON on all the three domains.
To demonstrate that Hi-Viscont is better for continual learning, we compare these models with no pre-trained concepts.
We present the mean and standard deviation of the F1 metric across the three datasets as our major results.
Each of these results is obtained from five trials with different splits of concepts and image.
We evaluate the question-answer pairs for all concepts for all the three domains on images that are not shown in the pre-train or the train phase. 
Images used for testing are never seen by the model in any phase of training for both train concepts and test concepts.
Additional statistics(precision, recall, and t-test) and a more detailed analysis can be found in the Appendix.

\begin{table}[H]
\resizebox{\columnwidth}{!}{%
\begin{tabular}{c|ccccc}
\toprule
Mtd.    & Species        & Genera         & Family         & Order          & Class          \\ 
\midrule
HV & 87.1$\pm$2.0 & \textbf{90.4$\pm$0.6} & \textbf{90.7$\pm$1.7} & \textbf{92.0$\pm$0.8} & 95.9$\pm$8.2           \\
FCN    & 86.5$\pm$1.4         & 88.2$\pm$1.0          & 84.3$\pm$1.4 & 84.3$\pm$3.2 & 99.3$\pm$1.0  \\ 
\bottomrule
\end{tabular}%
}
\caption{The average F1 score and standard deviation of Hi-Viscont (HV) and FALCON (FCN) on the CUB dataset by the depth of concepts in the hierarchy.}
\label{tab:CUB_depth}
\end{table}

\subsection{VQA Results}
In Table~\ref{tab:test_three}, we present the results on the VQA task for test concepts.
Our model, Hi-Viscont achieves comparable results to the baseline state-of-the-art FALCON model on test concepts in all three domains.
Given that in a concept network there are fewer parent concepts than leaf concepts, the performance of both models is comparable in such a general test case.
However, when we split the concepts by their depth in the hierarchy, Hi-Viscont shines and achieves a significantly better performance with the parental nodes, which will be discussed by each domain separately.

\noindent\textbf{CUB dataset:} 
We present our results for concepts by their level in the taxonomy in Table \ref{tab:CUB_depth}. 
Hi-Viscont is better with significance for concepts in the level of Genera($p<0.001$), Family($p=0.001$), and Order($p=0.001$) according to paired t-tests.
Species are the leaf level concepts where the models again perform comparably as expected. This is because the leaf level updates of Hi-Viscont and FALCON do not differ significantly.
As there is only one highest level ancestor for the Class with CUB there is no negative example for it in the dataset leading to similar performance by both models as the answer is always \texttt{True}.

\begin{table}[]
\resizebox{\columnwidth}{!}{%
\begin{tabular}{c|ccc}
\toprule
Method    & Object        & Color         & Affordance           \\
\midrule
Hi-Viscont &  88.46$\pm$1.58  & \textbf{99.24$\pm$0.70} & \textbf{89.86$\pm$9.12}          \\
FALCON    & 89.28$\pm$0.93         & 87.27$\pm$5.83          & 57.35$\pm$9.23 \\ 
\bottomrule
\end{tabular}%
}
\caption{The average F1 score and standard deviation of Hi-Viscont and FALCON on the house construction domain by type of concepts.}
\label{tab:house_type}
\end{table}

\noindent\textbf{House construction domain:} 
In this domain, the Color and  Affordance concepts are non-leaf nodes in the hierarchy, whereas the object concepts are the leaf nodes. 
Following expectations, as demonstrated in Table~\ref{tab:house_type}, Hi-Viscont has a comparable performance to FALCON in the leaf node object concepts, while achieving significant improvements in both Color ($p=0.005$) and Affordance (non-leaf) concepts ($p=0.002$) according to the pairwise t-tests.

\begin{table}[b]
\centering
\resizebox{0.75\columnwidth}{!}{%
\begin{tabular}{c|cc}
\toprule
Method          & Leaf       & Non-leaf           \\
\midrule
Hi-Viscont  & 87.93$\pm$3.40 & \textbf{85.84$\pm$5.79}          \\
FALCON          & 88.99$\pm$3.75         & 66.15$\pm$5.34 \\ 
\bottomrule
\end{tabular}%
}
\caption{The average F1 score and standard deviation of Hi-Viscont and FALCON on the zoo domain by type of concepts.}
\label{tab:zoo_type}
\end{table}

\noindent\textbf{Zoo Domain} 
In the zoo domain, leaf concepts are not at equivalent depths from the root node forcing us to analyze the performance crudely with respect to leaf and non-leaf nodes in Table~\ref{tab:zoo_type}.
Similarly, Hi-Viscont achieves a comparable performance at leaf level concepts, but becomes significantly better than FALCON in the non-leaf concepts ($p=0.001$).

\begin{table}[]
\resizebox{\columnwidth}{!}{%
\begin{tabular}{c|c|c}
\toprule

  Metrics                 & Hi-Viscont    & FALCON        \\ \midrule
Success Rate(\%)  & $50.00\pm51.45$ & $16.67\pm38.25$ \\ 
Node Accuracy(\%) & $81.25\pm21.11$ & $61.81\pm23.67$ \\ 
Comparative       & $5.44\pm2.68$   & $0.39\pm0.78$   \\ 
Trust              & $58.56\pm11.60$ & $51.94\pm13.91$ \\ 
SUS                & $46.72\pm10.33$ & $43.33\pm11.26$ \\ 
Intelligence       & $33.00\pm5.90$  & $28.61\pm7.37$  \\ 
Natural            & $13.89\pm4.19$  & $12.11\pm4.10$  \\ \bottomrule
\end{tabular}%
}
\caption{The results of the human-subjects experiment. Success Rate and Node Accuracy are measured in percentage. Hi-Viscont is better than FALCON on all metrics with significance as described in the Human Subjects Study Section.}
\label{tab:human_subject}
\end{table}
\subsection{Human Subjects Study}
We conducted a human-subjects study with $18$ participants ($22.22\%$ female, mean age = $25.36$, standard deviation = $3.49$). To design our study we conducted pilot studies with $10$ participants.
Each participant completed three phases of interaction with the robot on the house construction domain and filled out all surveys.
Our experiment results with respect to each hypothesis are as follows:

\noindent\textbf{Hypothesis 1: -} Hi-Viscont achieves a $33.33\%$ improvement in success rate(SR) compared to FALCON. Results from the Wilcoxon signed-rank test indicate that Hi-Viscont's SR is significantly better than FALCON ($Z=0.0$, $p=0.014$).

\noindent\textbf{Hypothesis 2: -} Hi-Viscont achieves a $19.44\%$ improvements in accuracy at node level compared to FALCON. Results from the Wilcoxon signed-rank test indicate that Hi-Viscont's node level accuracy is significantly better than FALCON($Z=1.5$, $p=0.005$).

\noindent\textbf{Hypothesis 3: -} Hi-Viscont achieves higher ratings on subjective metrics than FALCON. Users prefer Hi-Viscont in all the scales that we measured with significance: Trust($t=2.325$, $p=0.016$, $df=17$), SUS($t=2.428$, $p=0.013$, $df=17$), Perceived Intelligence($t=2.591$, $p=0.010$, $df=17$), and Anthropomorphism ($t=2.924$, $p=0.005$, $df=17$), suggested by paired t-test. Additionally, results from Wilcoxon signed-rank test suggest Hi-Viscont is significantly preferred over FALCON($Z=0.0$, $p<0.001$).

%% file: sections/discussions.tex
\section{Limitations}
There are a few clear limitations of our approach.
Firstly, although that we tested Hi-Vicont on a large VQA dataset, we conducted our robotics study of visual task learning only on the House domain, which contains a small number of objects. We would like to increase the task complexity and the number of objects available in the domain in the future. 
Secondly, the proposed method does not generalize across different domains automatically akin to a foundation model. Using this method on a completely new domain requires us to train the concept net model from scratch.
Thirdly, the interaction between users and the robots is controlled without being completely open and dynamic. Even though a fixed template for their language is not required, the users have to follow specific turn-taking rules. Lastly, our study uses college-age human subject's and we would like a wider sample of the population using our system.

\section{Conclusion}
In conclusion, we present Hi-Viscont, a novel concept learning framework that actively updates the representations of known concepts which is essential in continual learning settings such as robotics.
Hi-Viscont achieves comparable performance to SOTA FALCON model on VQA task across three domains in leaf level concepts, and is significantly better on non-leaf concepts.
Moreover, Hi-Viscont enables robots to learn a visual task from in-situ interactions by representing visual tasks with a scene graph. This approach allows zero-shot generalization to an unseen task of the same type.
Finally, we conducted a human-subjects experiment to demonstrate Hi-Viscont's ability to learn visual tasks and concepts from in-situ interactions from participants that have no domain knowledge in the real world.

%% file: appendices/training_details.tex
\begin{figure*}[]
  \centering
\begin{subfigure}{0.25\textwidth}
\includegraphics[width=\textwidth]{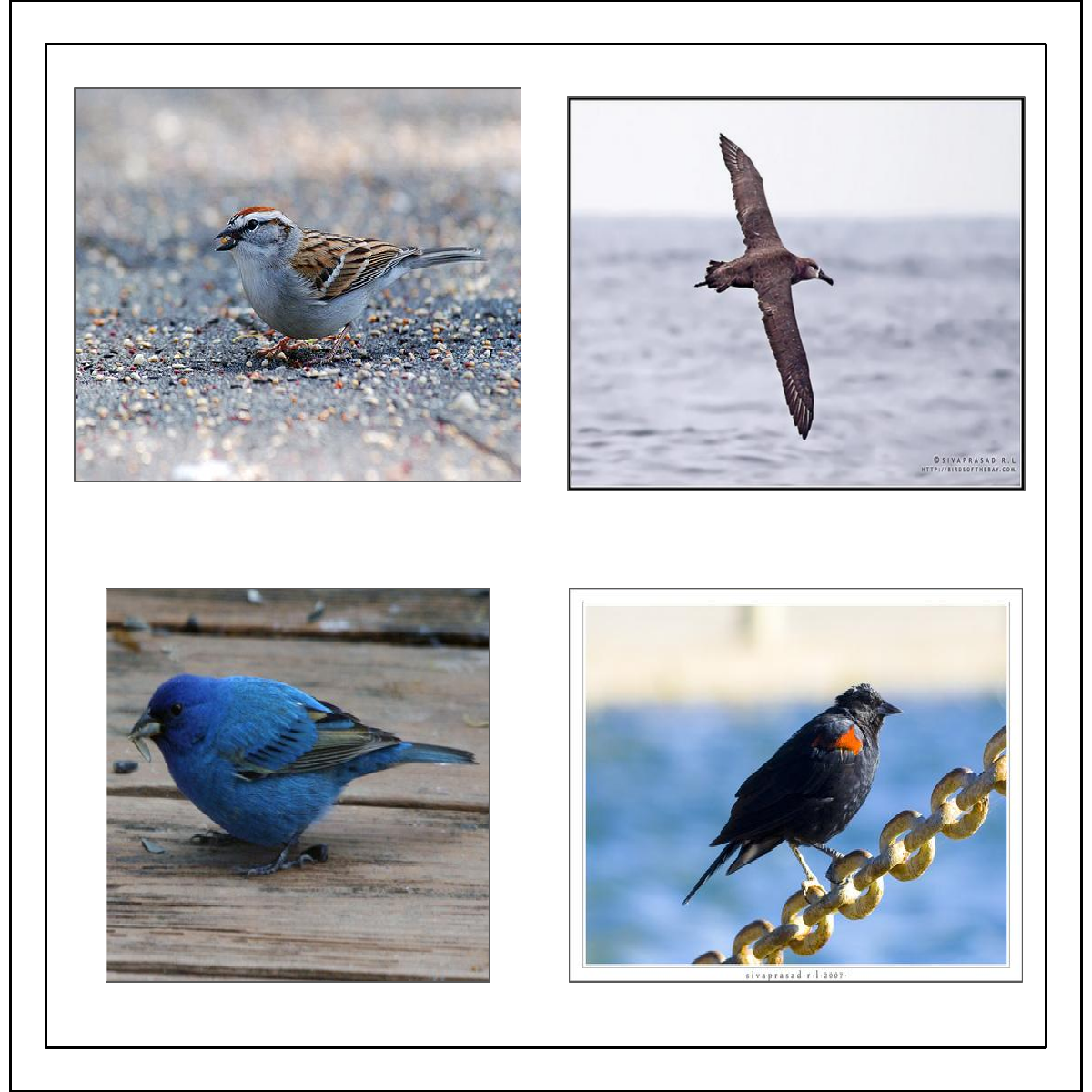}
  \subcaption{CUB dataset.}
\end{subfigure}
\hspace{0.04\textwidth}
\begin{subfigure}{0.25\textwidth}
\includegraphics[width=\textwidth]{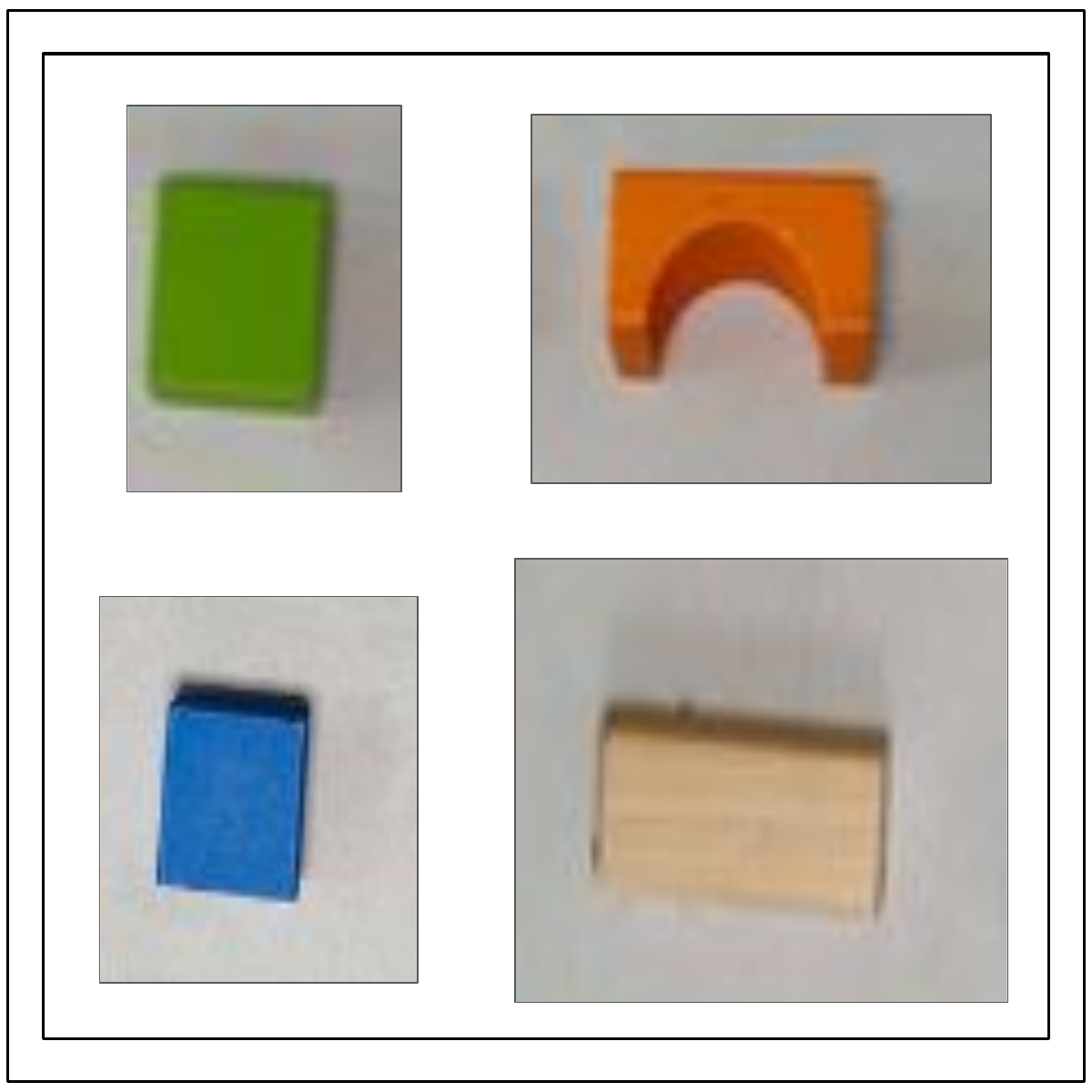}
  \subcaption{House construction domain}
\end{subfigure}
\hspace{0.04\textwidth}
\begin{subfigure}{0.25\textwidth}
\includegraphics[width=\textwidth]{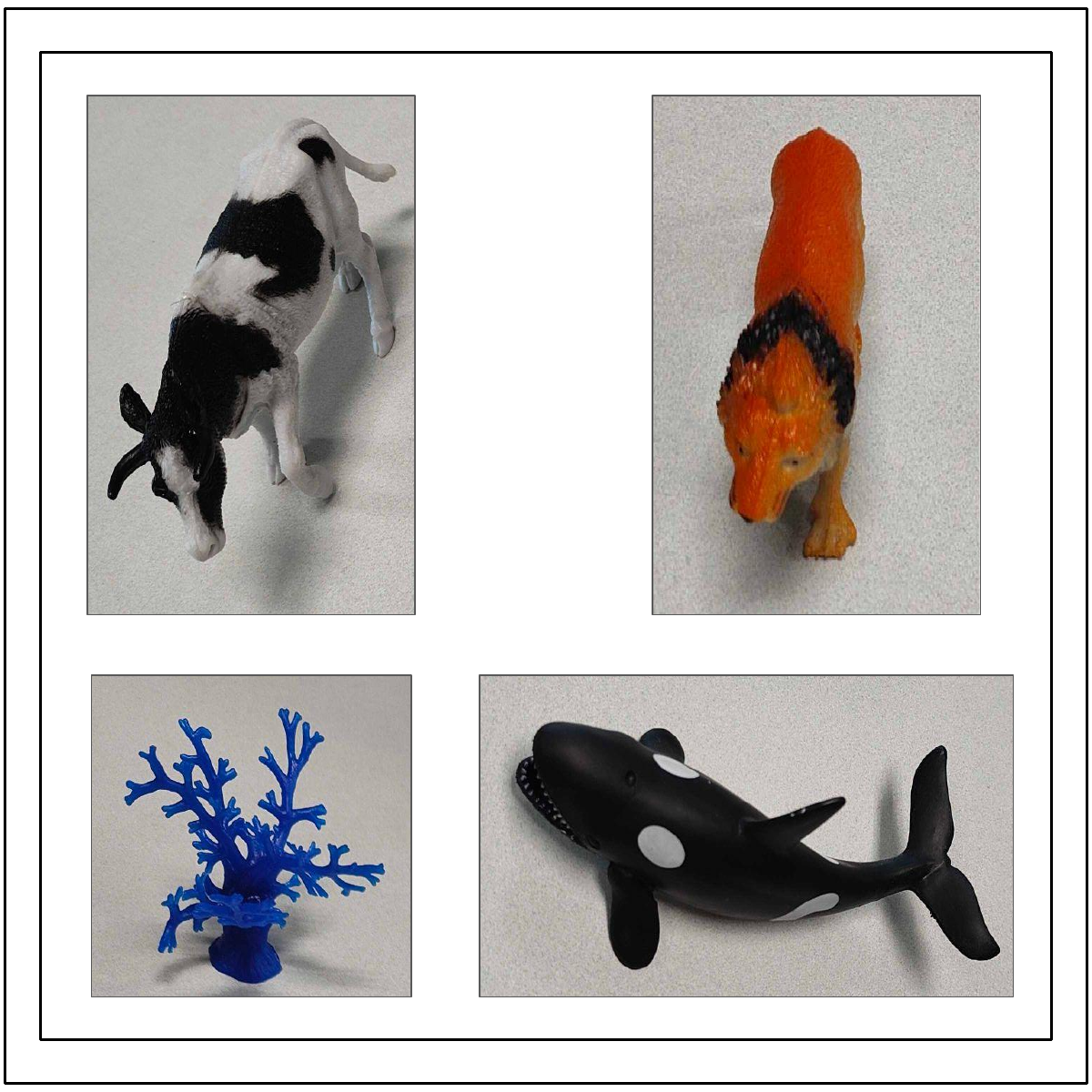}
  \subcaption{Zoo domain}
\end{subfigure}
\caption{Sample images from the three domains in this work.}
\label{fig:sampled_figures}
\end{figure*}



\subsection{Training Pipeline}
 We explain our training pipeline in this section. 
 The concepts from the dataset is divided into three groups: $C_{pretrain}, C_{train}$ and $C_{test}$, where the pre-train concepts $C_{pretrain}$ represent the pre-existing nodes in the knowledge graph. The training of the concept net model consists of three stages, the pre-training for the visual feature extractor, the pre-training for the embedding of pre-train concepts $C_{pretrain}$, and the training to update the knowledge graph with train concepts $C_{train}$. 
 \paragraph{Pre-training the Visual Feature Extractor.}
 In the first pre-training stage, we generate a VQA dataset on both the pre-train concepts $C_{pretrain}$ and the train concepts $C_{train}$. The purpose of this stage is to expose the visual feature extractor with a larger variation of visual features. 
 We jointly pre-train the visual feature extractor and the embedding for the pre-train concepts $C_{pretrain}$ and the train concepts $C_{train}$ with the visual question-answering task in this stage.
 After this pre-training stage, the embeddings of all pre-train concepts and train concepts will be discarded.
 \paragraph{Pre-training Pre-train Concepts.}
 The embedding of pre-train concepts $C_{pretrain}$ is obtained through gradient descent in this pre-train phase.
 For this phase, we generate a VQA dataset on the pre-train concepts $C_{pretrain}$ only.
 After we warmup the visual feature extractor in the first pre-training phase, we jointly train the visual feature extractor and the embedding for the pre-train concepts in this phase using the same VQA task.
 This pre-training step is skipped under the setting where the concept net has zero prior knowledge of the concepts, which is the setting of all of our experiments.
 \paragraph{Training.} 
 After we have pre-trained the visual feature extractor and the embedding for the pre-train concepts,  we train the concept learner module during the training stage.  We freeze the weights of the visual feature extractor at this stage because otherwise the embeddings for the pre-train concept will not be usable.  Because we hope to train ARGNN to update the embedding for known concepts with information from unseen instances, we have to reset the embedding for all the pre-train concepts and train concepts,$C_{pretrain}$ and $C_{train}$, after all the train concepts are inserted to the network.  After inserting all the concepts within the train set in the final round, we do not reset the embedding for the train concepts and insert the concepts in the test set $C_{test}$. 

\subsection{Training Configurations}
In this section we describe the training configuration of the experiments for all the three domains. 
During the training phase, the model completes one round of training if it finishes to insert all the concepts in the training set once. For simplicity, we unify the steps of training with rounds of insertion.
For all the experiment results we report in this work, we adopted the configuration where there is no pre-train concepts.
As a results, the second phase of pre-training is skipped for all the three datasets.
For all the three domains, we train our model for completing the concept graphs 100 rounds, and the number of concept insertions varies depending on the split of the concepts. We start the training with a learning rate of 0.001 and decrease the learning rate by a factor of 0.1 in every 25 rounds of completing the knowledge graph in the training stage.
For CUB-200-2011 dataset, we train our model for 50000 iterations with a batch size of 10. We use an Adam optimizer with learning rate of 0.0001 in the pre-training phase of the visual feature extractor.
For the house construction domain and the zoo domain, we train our model for 5000 iterations with a batch size of 10. We use an Adam optimizer with learning rate of 0.0001 in the pre-training stage of the visual feature extractor.

\subsection{Robot Setup}
We describe the details for camera calibration.
We need to calibrate cameras with respect to the FR3 base frame. 
We take multiple pictures in different configurations of the FR3 end-effector to which an acuro market is attached. This allows us to find a transformation matrix which converts the coordinates from the  camera frame to the robot base frame. The place scene camera is used to find the length of the object occupying the current node of the scene graph. 

We describe how we compute the placement location for each object in detailed.
SAM is used to segment the objects placed in the place scene and find the bounding boxes of each placed object which are also the nodes of our scene graph.
This allows us to calculate the position of the next object by finding the relative position of the next node with respect to the current object being placed.
, referencing the position of the node of the scene, and calculating the length of the bounding box of the referenced node. 
we use a formula of shift = 1/2*max(bounding box of the referenced node length)+50 pixel space \textbf{Next node position}= Relation to the reference node(Reference node position,shift). The function relation to the reference node adds a shift to the reference node position based on its relation to the next node. For example, it adds the shift only to the x coordinate if there is "to the top of" relation, or in the case of "to the right of" relation, it adds only the y coordinate of the current position. In our scene graph, we are able to identify "to the top of ", "to the bottom of"," to the right of"," to the left of", "to the top right of"," to the top left of"," to the bottom right of", and "to the bottom left of" relations.\\
The Segment Anything Model is capable of separating the foreground from the background. This allows us to find the table mask and the segment of each object placed in the camera frame on the table. \\
The flow of our pipeline requires us to first demonstrate the visual scene with all the objects placed in the Task scene to make a structure with linguistic inputs. 
We have to make sure that the objects are placed at a distance that allows SAM to create separate segment boxes for the objects. Then we pass each segmented object to either FALCON or Hi-Viscont classifiers to classify conditioned on the given language query. The robot then picks the object with simple visuo-motor servoing. 
by the node information of the scene graph. 
Once we find the object to be picked we then calculate the center of the bounding box of that object and convert it to the Robot frame with the help of the transformation matrix. 
If in the process there is an incomplete or erroneous grasp, we reattempt the whole classification again autonomously. Once the object is grasped we then place the object into the Task scene, with the position calculated relatively 
with respect to the previously placed object nodes or the ground. This process is iteratively done until we have completed the whole scene graph.

%% file: appendices/dataset_statistics.tex


\begin{figure*}[h]
  \centering
\begin{subfigure}{0.28\textwidth}
\includegraphics[width=\textwidth]{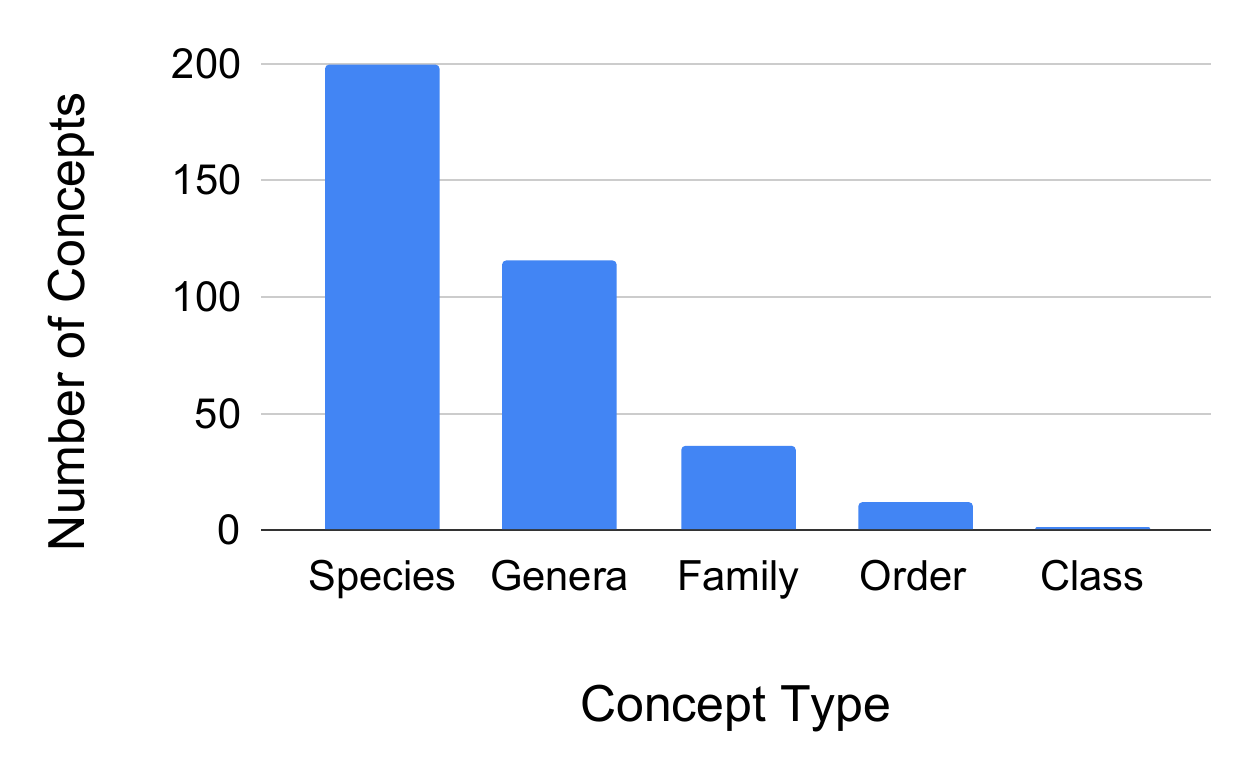}
  \subcaption{CUB-200-2011}
\end{subfigure}
\hspace{0.01\textwidth}
\begin{subfigure}{0.28\textwidth}
\includegraphics[width=\textwidth]{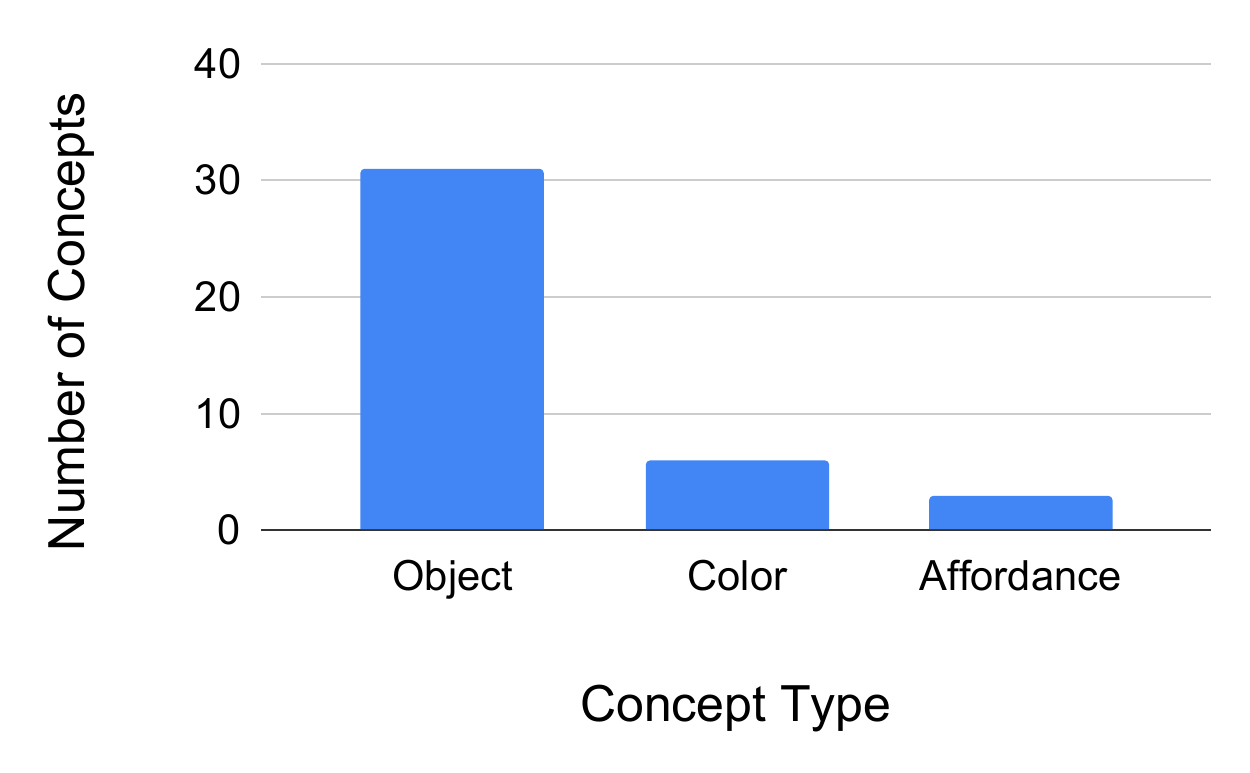}
  \subcaption{House Construction Domain}
\end{subfigure}
\hspace{0.01\textwidth}
\begin{subfigure}{0.28\textwidth}
\includegraphics[width=\textwidth]{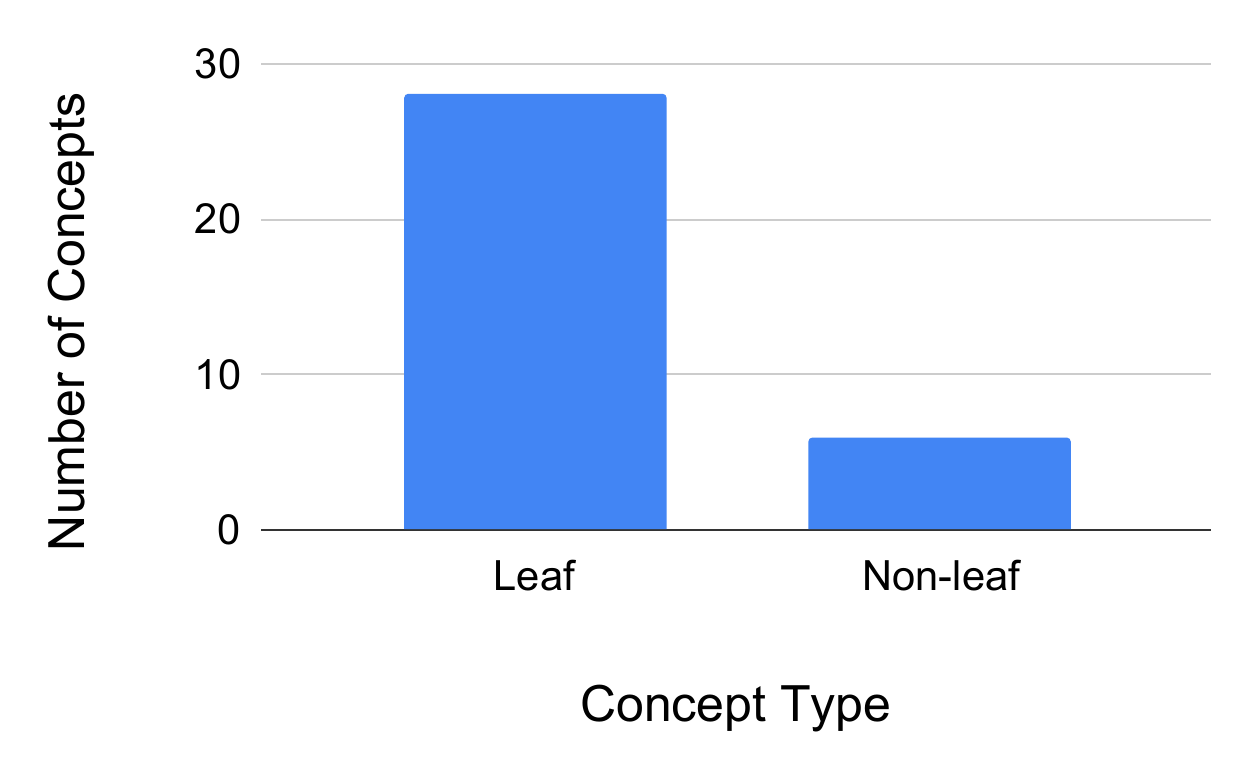}
  \subcaption{Zoo Domain}
\end{subfigure}

\caption{The number of concepts of each type for all the three domains.}
\label{fig:concept_type}
\end{figure*}

\begin{table*}[t]
\resizebox{\textwidth}{!}{%
\begin{tabular}{c|ccc|ccc}
\toprule
    & Train Image & Test Image & Total Image& Train Concept & Test Concept & Total Concept\\
\midrule
CUB-200-2011 & $8232$ & $3556$ & $11788$ & $307.8$ & $58.2$ & $365$  \\
House& $217$ & $93$ & $310$& $34$ & $6$ & $40$\\
Zoo& $196$ & $84$ & $280$& $28$ & $6$ & $34$\\
\bottomrule
\end{tabular}%
}
\caption{Statistics for the three domains. The number of train and test concepts of the CUB-200-2011 dataset comes from the average of five different splits, resulted in decimal numbers.}
\label{tab:all_stats}
\end{table*}

We split our dataset at two different levels:- the level of concepts and the level of images. The images are divided into $70\%$ for training and $30\%$ for testing, and the testing images are used to evaluate for both seen and unseen concepts. At the concept level, we use five-fold validation. We train the model with $80\%$ of concepts and test on $20\%$ while covering all folds. Because the split of the non-leaf concepts need to guarantee that leaf concepts are unseen as well, our pick of the folds depends on the structure of the concept graph and is not as random as the regular five-fold validation.
The detailed descriptions of each domain are as follows:

\noindent\textbf{CUB-200-2011 dataset}\cite{WahCUB_200_2011} is a standard dataset to demonstrate visual concept learning.
It contains $11,788$ images for $200$ bird classes. Using the following  bird taxonomy\cite{Sullivan2009eBirdAC}, we added the hypernyms of the bird classes and used the bird taxonomy as the hierarchy of concepts.
Following the design of the dense graph propagation~\cite{kampffmeyer2019rethinking}, the relation of each concept includes all of its ancestors.

\noindent\textbf{The house construction domain} includes $31$ types of building block objects.
Each object has $10$ different images. 
To introduce relations between concepts, we additionally add $6$ different concepts and $3$ different affordances of objects.
The dataset on the house construction domain includes $310$ images and $40$ concepts in total.
This domain is also used to train the models for the human subject study.

\noindent\textbf{The zoo domain} includes $28$ different types of objects. 
Similar to the house domain, we add $6$ general concepts to introduce a hierarchy for the concepts.
The dataset on the zoo domain includes $280$ images and $34$ concepts in total.

Additionally, the detailed statistics of image and concept of each split of each domain are presented in Table~\ref{tab:all_stats}, and the statistics of concept type are presented in Fig.~\ref{fig:concept_type}. Fig.~\ref{fig:sampled_figures} exhibits some sample images from the three domains.

%% file: appendices/detailed_results.tex
\begin{table*}[t]
\resizebox{\textwidth}{!}{%
\begin{tabular}{c|cc|cc|cc}
\toprule
    & CUB-P & CUB-R & House-P& House-R & Zoo-P & Zoo-R \\
\midrule
Hi-Viscont & $91.03 \pm 3.00$ & $63.35\pm 9.69$ & $87.39\pm 1.71$  & $85.89\pm 9.95$ & $85.24\pm 5.41$ & $83.37\pm 15.67$\\
FALCON & $90.60\pm 4.18$  & $62.15\pm 8.38$  & $87.62\pm 1.71$ & $87.04\pm 8.29$ & $84.10\pm 5.86$& $87.41\pm 12.98$\\
\bottomrule
\end{tabular}%
}
\caption{The mean and standard deviation of precision and recall of Hi-Viscont and FALCON on the test concepts across all the three domains.}
\label{tab:detailed_test}
\end{table*}

\begin{table*}[t]
\resizebox{\textwidth}{!}{%
\begin{tabular}{c|cc|cc|cc|cc|cc}
\toprule
    &  Species-P& Species-R & Genera-P & Genera-R & Family-P & Family-R & Order-P & Order-R & Class-P & Class-R\\
\midrule
Hi-Viscont & $98.32\pm 0.63$  & $78.20\pm 3.56$ & $97.06\pm 0.44$ & $84.61\pm 0.90$ & $93.51\pm 2.33$& $88.12\pm 2.08$& $94.68\pm 1.58$& $89.40\pm 1.68$& $1\pm 0.00$ & $93.04\pm 13.71$\\
FALCON & $98.28\pm 0.59$ & $77.35\pm 2.59$& $96.66\pm 0.86$& $81.17\pm 1.24$ & $87.55\pm 3.13$& $81.43\pm 2.40$& $85.70\pm 3.70$& $83.10\pm 5.28$& $1\pm 0.00$ & $98.63\pm 1.85$ \\
\bottomrule
\end{tabular}%
}
\caption{The mean and standard deviation of precision and recall of the two models on CUB-200-2011 dataset by depth of concepts.}
\label{tab:detailed_CUB}
\end{table*}

\begin{table*}[t]
\resizebox{\textwidth}{!}{%
\begin{tabular}{c|cc|cc|cc}
\toprule
    & Object-P & Object-R & Color-P& Color-R & Affordance-P & Affordance-R \\
\midrule
Hi-Viscont & $86.19\pm 1.81$& $90.91\pm 3.08$& $99.94\pm 0.09$& $98.56\pm 1.41$& $95.51\pm 1.34$& $85.89\pm 14.86$\\
FALCON &$87.16\pm 1.43$& $91.57\pm 2.66$& $91.71\pm 2.28$& $83.80\pm 10.07$& $91.37\pm 1.71$& $42.25\pm 9.32$\\
\bottomrule
\end{tabular}%
}
\caption{The mean and standard deviation of precision and recall of the two models on the house domain by type of concepts.}
\label{tab:detailed_house}
\end{table*}

\begin{table*}[t]
\resizebox{\textwidth}{!}{%
\begin{tabular}{c|cc|cc}
\toprule
    & Leaf-P & Leaf-R & Non-leaf-P& Non-Leaf-R\\
\midrule
Hi-Viscont &$82.97\pm 5.32$& $93.89\pm 5.25$&$77.30\pm 7.77$& $96.93 \pm 5.97$\\
FALCON & $84.01\pm 5.29$& $94.90\pm 5.03$& $76.80\pm 4.08$& $58.38\pm 7.09$\\
\bottomrule
\end{tabular}%
}
\caption{The mean and standard deviation of precision and recall for the two models on the zoo domain by type of concepts.}
\label{tab:detailed_zoo}
\end{table*}

\subsection{Detailed Results on VQA}
We present the detailed results and analysis on VQA for each domain. Table~\ref{tab:detailed_test} presents the mean and standard deviation of the precision and recall of Hi-Viscont and FALCON on test concepts across all the three domains. The performance of both model is comparable on the test concepts on both precision and recall because majority of the test concepts are leaf concepts. Table~\ref{tab:detailed_CUB}, Table~\ref{tab:detailed_house}, and Table~\ref{tab:detailed_zoo} present the mean and standard deviation of the precision and recall for the CUB-200-2011 dataset, the house domain, and the zoo domain by the type of concept repectively. For all the three domains, Hi-Viscont consistently outperform FALCON on the recall metric for non-leaf concepts with a large margin because of the active updates from the ARGNN. Such results further demonstrate the importance of the information propagation from children to ancestor nodes in a continual learning setting.

\subsection{Statistic Tests on VQA}
In this section, we presents the statistic tests between Hi-Viscont and FALCON for all the three domains. 
\paragraph{CUB-200-2011.} Results of paired t-test suggest that Hi-Viscont achieves higher F1 scores with significance for concepts in Genera($p<0.001$), Family($p<0.001$), and Order($p=0.005$).
\paragraph{House Construction Domain.} Results of paired t-test suggests that Hi-Viscont achieves higher F1 scores with significance for color concepts($p=0.005$) and affordance concepts($p=0.002$).
\paragraph{Zoo Domain.} Results of paired t-test suggests that Hi-Viscont achieves higher F1 scores with significance for non-leaf concepts($p=0.001$).


\subsection{Detailed Results on Human Subject Experiments}
We present the results of statistical tests for each scale we measure for the human subject study in detailed in this subsection. 
For each metric, we report the results of a Shapiro-Wilk Normality test. If the data from such metric passes the normality test($p>0.05$), we report the results of a paired t-test.
Otherwise, we report the results of Wilcoxon signed rank test for that metric. 
\paragraph{Trust.} Results from Shapiro Wilk test suggest that our data in the Trust metric satisfies the condition for a parametric test($W=0.960, p=0.599$). Results from paired t-test suggest that Hi-Viscont is considered better than FALCON by users in the Trust metric with significance($t=2.325, p=0.016, df=17$).
\paragraph{SUS.} Results from Shapiro Wilk test suggest that our data in the SUS metric satisfies the condition for a parametric test($W=0.978, p=0.925$). Results from paired t-test suggest that Hi-Viscont is considered better than FALCON by users in the SUS metric with significance($t=2.428, p=0.013, df=17$).
\paragraph{Intelligence.}Results from Shapiro Wilk test suggest that our data in the Intelligence metric satisfies the condition for a parametric test($W=0.956, p=0.520$). Results from paired t-test suggest that Hi-Viscont is considered better than FALCON by users in the Intelligence metric with significance($t=2.591, p=0.010, df=17$).
\paragraph{Natural.}Results from Shapiro Wilk test suggest that our data in the Anthropomorphism metric satisfies the condition for a parametric test($W=0.937, p=0.255$). Results from paired t-test suggest that Hi-Viscont is considered better than FALCON by users in the Anthropomorphism metric with significance($t=2.924, p=0.005, df=17$).
\paragraph{Comparative.}Conditions for normality were not met for the data points to run a t-test. Hence we conducted a Wilcoxon Signed-Rank test to compare Hi-Viscont and FALCON in the comparative metric. Results from Wilcoxon signed-rank test suggest that Hi-Viscont is preferred by user in the direct comparison with FALCON with significance($Z=2.0, p<0.001$).
\paragraph{Success Rate.}Conditions for normality were not met for the data points to run a t-test. Hence we conducted a Wilcoxon Signed-Rank test to compare Hi-Viscont and FALCON in success rate. Results from Wilcoxon signed-rank test suggest that Hi-Viscont is better than FALCON in success rate with significance($Z=0.0, p=0.014$).
\paragraph{Node-level Accuracy.}Conditions for normality were not met for the data points to run a t-test. Hence we conducted a Wilcoxon Signed-Rank test to compare Hi-Viscont and FALCON in node-level accuracy. Results from Wilcoxon signed-rank test suggest that Hi-Viscont is better than FALCON in node-level accuracy with significance($Z=1.5, p=0.005$).